\documentclass[10pt,twocolumn,letterpaper]{article}

\usepackage[pagenumbers]{cvpr} 

\usepackage{multicol}
\usepackage{makecell}
\usepackage{multirow}
\usepackage{booktabs}
\usepackage[dvipsnames]{xcolor}

\definecolor{cvprblue}{rgb}{0.21,0.49,0.74}
\usepackage[pagebackref,breaklinks,colorlinks,citecolor=cvprblue]{hyperref}

\def\paperID{22} 
\def\confName{CVPR}
\def\confYear{2024}

\title{Run-time Monitoring of 3D Object Detection in Automated Driving Systems Using Early Layer Neural Activation Patterns}

\author{Hakan Yekta Yatbaz$^1$\hspace{0.4cm}  Mehrdad Dianati$^{1,2}$ \hspace{0.4cm}Konstantinos Koufos$^1$\hspace{0.4cm}Roger Woodman$^1$\\
$^1$WMG, University of Warwick\\
$^2$School of Electronics, Electrical Engineering and Computer Science (EEECS)\\
Queen’s University of Belfast \\
{\tt\small (hakan.yatbaz, m.dianati, konstantinos.koufos, r.woodman)@warwick.ac.uk,\vspace{0.2cm}   m.dianati@qub.ac.uk}
}

\begin{document}
\maketitle
\begin{abstract}
Monitoring the integrity of object detection for errors within the perception module of automated driving systems (ADS) is paramount for ensuring safety. Despite recent advancements in deep neural network (DNN)-based object detectors, their susceptibility to detection errors, particularly in the less-explored realm of 3D object detection, remains a significant concern. State-of-the-art integrity monitoring (also known as introspection) mechanisms in 2D object detection mainly utilise the activation patterns in the final layer of the DNN-based detector's backbone. However, that may not sufficiently address the complexities and sparsity of data in 3D object detection. To this end, we conduct, in this article, an extensive investigation into the effects of activation patterns extracted from various layers of the backbone network for introspecting the operation of 3D object detectors. Through a comparative analysis using Kitti and NuScenes datasets with PointPillars and CenterPoint detectors, we demonstrate that using earlier layers' activation patterns enhances the error detection performance of the integrity monitoring system, yet increases computational complexity. To address the real-time operation requirements in ADS, we also introduce a novel introspection method that combines activation patterns from multiple layers of the detector's backbone and report its performance. 
\end{abstract}    
\section{Introduction}
Effective and faithful perception of the surroundings is crucial for automated driving systems (ADS), as failure to capture the road traffic conditions can pose serious safety concerns, possibly leading to incidents that may involve fatalities or severe injuries~\cite{eduardo1, eduardo2}. This highlights the need for perception systems which can robustly handle runtime errors, necessitating continuous monitoring mechanisms of the perception system's integrity~\cite{hakansurvey, rahmansurvey}. Once a perception error is detected, the integrity monitoring systems issue an alert that can trigger a driver takeover in Level 3 or execution of a Minimum Risk Manoeuvre in Level 4 ADS.

\begin{figure}[t]
    \centering
    \includegraphics[width =\linewidth]{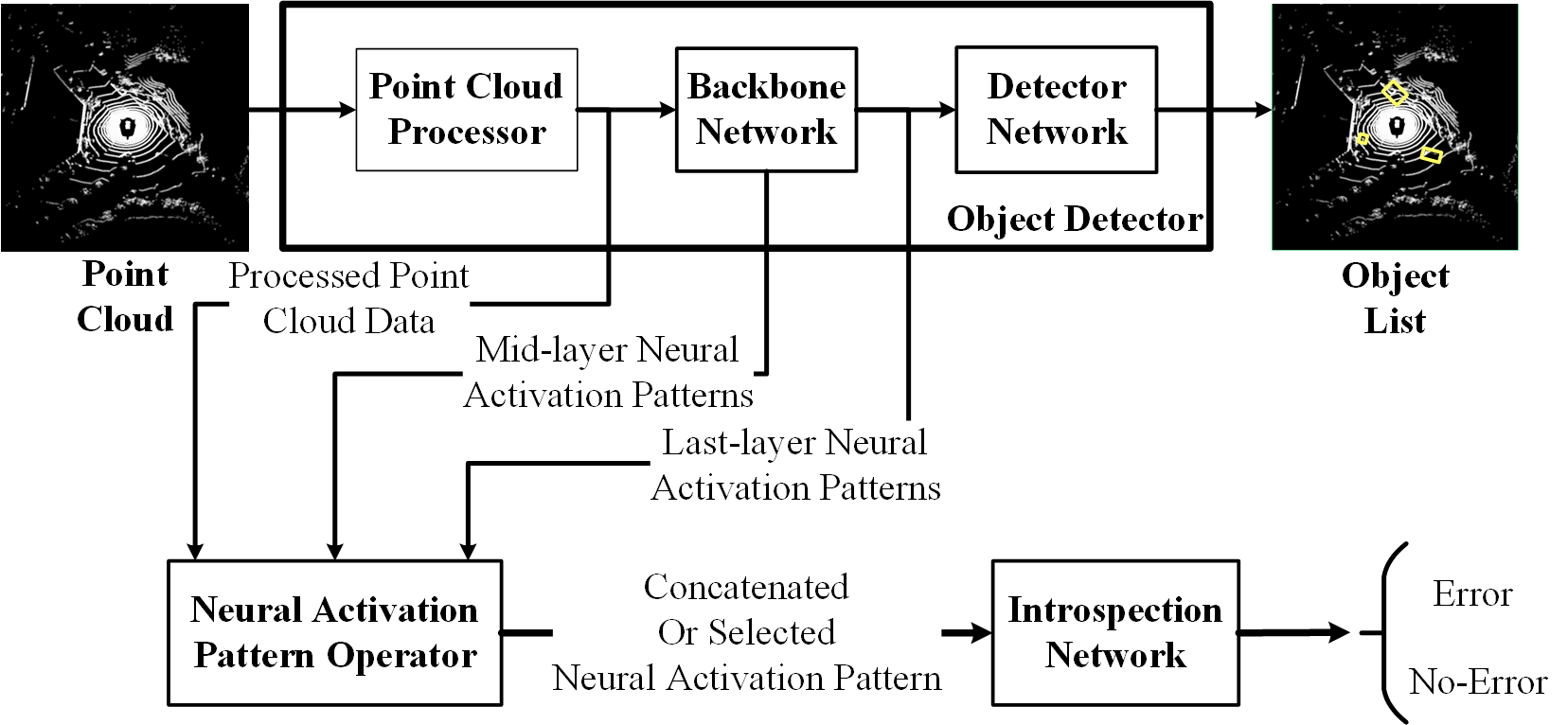}
    \caption{LiDAR-based object detection pipeline depicted at the top starting with a processor network that extracts features from the point cloud. The extracted features are then processed by a backbone network to compute neural activation patterns at the mid and last layers. These patterns are managed by the Neural Activation Pattern Operator as part of the introspection framework, which either combines them (our proposed method) or chooses patterns from an earlier layer (as part of our investigation). Finally, the selected pattern feeds into the Introspection Network, which classifies the collected point cloud as `Error' or `No-Error'.}
    \label{fig:overview}
\end{figure}

State-of-the-art (SOTA) object detection mechanisms that leverage various types of deep neural networks (DNNs) have recently shown remarkable performance on several benchmarks and continue to improve~\cite{detectionsurvey,detectionsurvey2}. Despite these advancements, object detectors in ADS are not infallible, especially in the complex and dynamic driving domain, due to the broad range and complexity of driving scenarios involving several road users as well as due to the sensitivity of perception sensors to various impairments, noise, occlusions and faults~\cite{Eduardo2020, Buchholz}. A common approach to enhance safety and trust in ADS is the deployment of runtime monitoring or ``introspection" mechanisms that continuously assess the integrity of perception outputs.


Despite gaining significant attention in the past decade, introspection of DNN-based mechanisms for object detection in ADS has so far focused on camera-based 2D object detection and classification~\cite{hakansurvey}. As ADS aspire to achieve higher levels of autonomy, they increasingly incorporate additional sensors, such as LiDARs, for robust object detection in adverse and variable conditions. Moreover, the inherently three-dimensional nature of the world requires a comprehensive understanding of the environment in 3D to ensure the resilience and robustness of ADS applications. Despite that, only limited attempts have been made so far for the introspection of LiDAR-based 3D object detection. The existing works in the literature are mostly implemented based on the monitoring of the confidence levels at the output of the detector~\cite{feng2018towards,feng2019can}.

A new method for introspecting the performance of camera-based 2D object detection has been recently proposed that leverages the activation maps at the output of the backbone network in conjunction with the mAP score~\cite{hakantiv,hakaniccv}. Specifically, the introspection network takes the latent features as input and is trained to declare detection errors once the predicted mAP is less than a threshold. This method has gained popularity due to its flexibility, ease of integration into other systems, and superior performance in comparison with SOTA models for introspecting the detections of YOLOv8 and Faster R-CNN. That motivates the investigation of utilising neural activation patterns for the introspection of LiDAR-based point clouds, which is what we will do in this paper.

This article evaluates introspection mechanisms using neural activation patterns of the DNN-based 3D object detectors that use LiDAR-based point clouds, investigating the effectiveness of early layer activations as opposed to the traditional reliance on activations from the final layer. This is based on the hypothesis that the sparse nature of LiDAR data, unlike camera-based 2D detection, may render final layer activations inadequate for accurately detecting failures in object detection. Additionally, we introduce a novel approach that concatenates neural activation patterns from multiple layers for better error detection. For this purpose, we utilise the widely used baseline models in ADS, PointPillars~\cite{pointpillars} and CenterPoint~\cite{centerpoint}, and extract activation patterns from multiple layers of its backbone model, SECOND~\cite{SECOND}. An error is declared if the object detector misses at least one actor, and the (binary) labels (`1' for Error and `0' for No-Error) are paired with the concatenated neural activation patterns. Subsequently, an introspector convolutional neural network (CNN) is trained on the generated pairs, hereafter referred to as the error dataset. A high-level summary of our mechanism is presented in Fig.~\ref{fig:overview}.
In summary, the contributions of this paper are:
\begin{itemize}
    \item A novel introspection approach for 3D object detection is designed to integrate neural activation patterns from multiple layers of the backbone network. 
    \item We investigate the efficacy of extracting neural activation patterns from earlier layers versus traditional reliance on the final layer, especially in the context of LiDAR's sparse data nature as compared to camera-based 2D detection.    
    \item A qualitative and quantitative evaluation of the considered adapted introspection mechanisms is presented in terms of error detection capability on two well-known public driving datasets, Kitti~\cite{kitti} and NuScenes~\cite{nuscenes}, and widely used 3D object detector baselines in ADS, PointPillars~\cite{pointpillars} and CenterPoint~\cite{centerpoint}.
    \item The confidence distribution in the decisions of the considered introspection models is examined to gain a deeper understanding of their performance comparison. Their computational complexity is also assessed, because of the stringent real-time operation requirements in ADS.
\end{itemize}

The remainder of this paper is organised as follows: Section~\ref{sec:litr} examines existing research on object detection introspection. Section~\ref{sec:meth} details the proposed introspection method and adapted mechanism. Section~\ref{sec:perfeval} describes the experimental setup and assesses the performance. Section~\ref{sec:conc} concludes with the paper's main findings.

\section{Related Work}\label{sec:litr}
\label{sec:method}
\begin{figure*}
    \centering
    \includegraphics[width=0.7\linewidth]{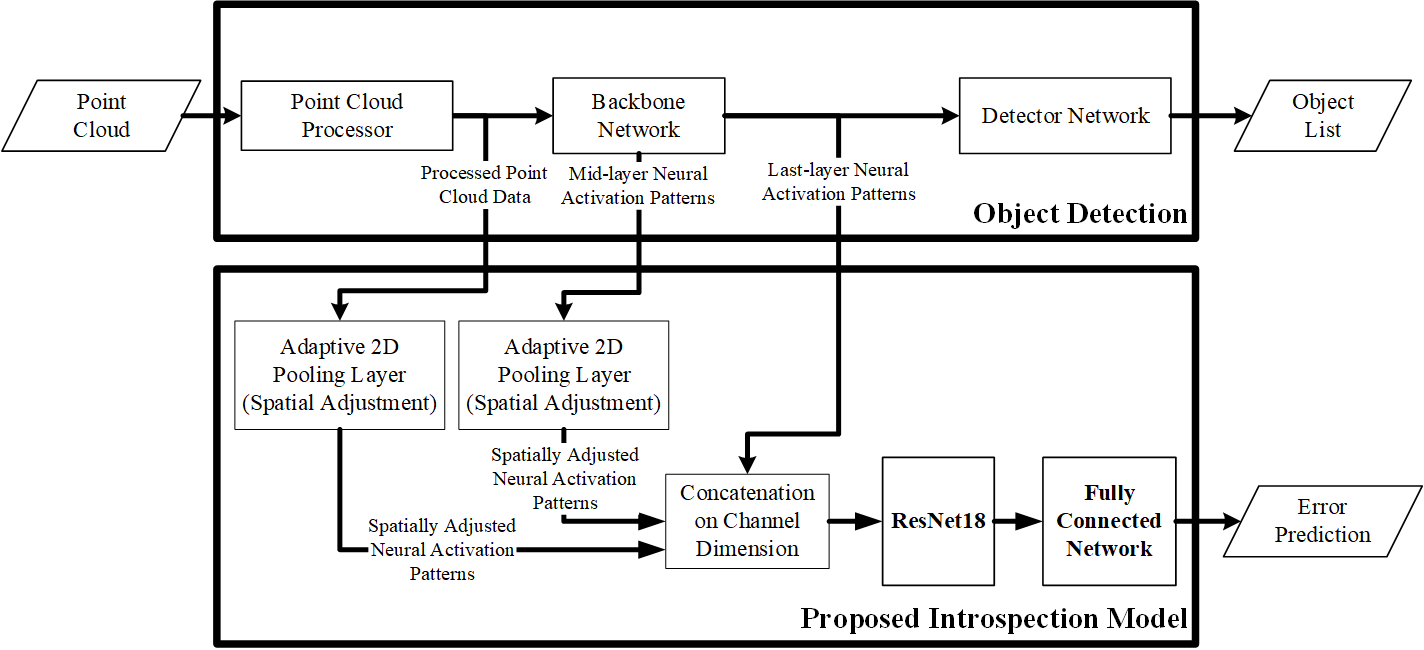} 
    \caption{Proposed introspection mechanism for LiDAR-based 3D object detection. The introspection mechanism depicted at the bottom captures the processed point cloud data, mid-layer neural activations, and backbone network outputs from the main object detection pipeline. An adaptive average pooling layer spatially adjusts these inputs to ensure uniform feature representation before concatenation, albeit with some resolution loss. The concatenated features are fed into the introspector network that comprises a ResNet18 for feature extraction and a fully-connected network for error prediction, ultimately assessing object detection errors as binary classification.}
    \label{fig:proposedframework}
\end{figure*}
This section offers a concise review of introspection methods in ADS for object detection.  Introspection mechanisms are categorised into confidence-based, performance-based, inconsistency-based, and past experience-based methods~\cite{hakansurvey}. The majority of the SOTA is concerned with introspecting camera-based 2D object detection, while there are only a few studies focusing on 3D object detection. Those studies model or estimate realistic confidence and uncertainty values at the object level, concerning more on the probabilistic nature of the models rather than the safety of the overall system. Hence, what we do in this paper constitutes a novel contribution to the SOTA introspection of LiDAR-based 3D object detection. 

\textbf{Confidence-based Introspection}: This category contains the studies modelling the uncertainty/confidence in object detection \cite{fengprobsurvey}. Key studies include testing various confidence mechanisms on point cloud datasets \cite{9922026,feng2018towards,feng2019can,feng3,xie3d}, and Cen \textit{et al.}'s unsupervised clustering approach for open-set 3D object detection, where uncertainty is quantified using Euclidean distances from raw points to class-specific `prototypes' \cite{9665821}.

\textbf{Performance-based Introspection}: This method focuses on detecting performance drops in object detection metrics, where mAP is the common metric used. For instance, the authors in~\cite{rahmanper} employed global pooling to extract features like mean and standard deviation from CNN outputs, predicting mAP drops for error detection. The study in~\cite{rahmancascade} enhanced this approach with temporal information and a cascaded neural network. In \cite{hakaniccv}, authors proposed an activation processing mechanism first introduced for out-of-distribution detection in classification task for error detection of 2D object detection. The studies in~\cite{rahmansign} and~\cite{posthoc2022} explored false negative detection using activation maps and handcrafted features, respectively. Yang \textit{et al.} in~\cite{fnyang} introduced a method predicting object-level false negatives independently of the underlying object detector.

\textbf{Inconsistency-based Introspection}: This method exploits multi-modality in ADS perception technologies. Examples include~\cite{failingtolearn}, which checked consistency between 2D object detectors and trackers using stereo and temporal cues, and~\cite{kdiagnose}, where Antonante~\textit{et al.} developed a diagnostic graph model for fault identification, further refined with a graph neural network in~\cite{antonante2022monitoring}.

\textbf{Past Experience-based Introspection}: That method involves creating a knowledge base with performance indicators and environmental characteristics. Studies in this category commonly evaluated the performance of ADS operating in a limited or the same environment such as ring buses.  Notable studies include a location-specific introspection approach in~\cite{fitforpurpose}, allowing robot autonomy only in reliably localised areas, and~\cite{learnfromexp}, which extended this approach by incorporating visual similarity-based experience.
\section{Method}\label{sec:meth}

This section introduces a novel introspection mechanism for 3D object detection on a per-frame basis, leveraging extracted activation maps from various stages of the object detector's backbone network, as well as the processed point cloud data. Unlike existing introspection studies on 2D object detection that predominantly focus on latent activation patterns for error identification, our approach addresses the unique challenges of 3D object detection with point cloud data. Point clouds, characterised by inherent sparsity, present significant challenges in 3D object detection tasks, differing from the dense information present in images. Furthermore, while 2D introspection models, such as those in~\cite{hakaniccv, hakantiv}, learn the relationship between activation patterns and mAP, the authors in~\cite{hakantiv} have highlighted that mAP can be misleading when there are different classes of objects present in the frame. For example, if a scene consists of multiple vehicles and a single pedestrian, where the pedestrian and majority of the vehicles are detected, but few vehicles are missed, the frame can still be labelled as no-error regardless of the missed vehicles' location. Hence, in this paper, we opt to identify the relationship between activation patterns and missed objects (false negatives) from the 3D point cloud to provide better safety, i.e., if at least one object is not detected, the frame is classified as `Error'.  


Acknowledging the above-mentioned limitations, 
the proposed mechanism employs an early-level fusion strategy, concatenating early-layer, mid-layer and output activation patterns from the backbone network into the introspection model. Including earlier and middle layers that have less fine-tuned activation patterns than the latent features aims at enhancing the introspector's ability to discern patterns indicative of missed objects. 
Consequently, it provides a robust introspective analysis, yielding a better understanding of the data and the intricacies of the neural network's processing without increasing the computational complexity and, hence, without compromising the model's real-time performance. 

\Cref{fig:proposedframework} presents the framework for introspection and the investigations tailored for LiDAR-based 3D object detection. The object detector module uses point cloud data, which undergoes processing in the ``Point Cloud Processor", ``Backbone Network", and ``Detector Network," depending on the type of the detector. During this process, our framework extracts the processed point cloud data, a middle layer neural network activation from the backbone network of the detector, and the output of the backbone network activations. The data extracted from the point cloud processor and mid-layers undergoes spatial adjustment by an adaptive average pooling layer, which is essential for maintaining consistent feature representation and concatenation. However, it is important to highlight that due to the reduction in resolution, some information is lost at this point.  Once the adjustment is finished, concatenation occurs on the channel dimension, ensuring that the comprehensive feature representation maintains spatial coherence.
The concatenated activation patterns are then processed by the introspection network, which follows the same architecture presented in~\cite{hakaniccv, hakantiv}. It utilises a ResNet18 network, which acts as a feature extractor to distil essential attributes for error prediction. The output of ResNet18 feeds into a fully connected network for error prediction. This is where the model determines whether the object detection system has failed to detect objects in the given frame. It is important to highlight that despite the proposed concatenation of activations, our framework is able to feed and train using activation maps from different layers individually. 

\section{Performance Evaluations}\label{sec:perfeval}
This section presents the experimental setup and performance evaluation of our study comparing introspection mechanisms for 3D object detection in ADS utilising neural activation patterns from different layers. Before that, we justify the selection of object detectors, driving datasets, adapted SOTA introspection mechanisms, and key performance indicators. 

\subsection{Object Detectors}
We investigate the behaviour of introspection systems on 3D object detection using two popular models. First, we utilise PointPillars~\cite{pointpillars}, a widely used baseline model in 3D object detection in ADS~\cite{autowareauto}. PointPillars proposes a novel encoder architecture, transforming the irregular and sparse 3D point clouds generated by LiDAR sensors into a structured format called ``pillars". These pillars are essentially vertical columns that capture the points in a defined columnar space, simplifying the complexity of 3D data processing. Once the data is organised into pillars, PointPillars employs a neural network to learn distinctive features from each pillar. The model projects these learned features onto a pseudo-image, enabling the use of a 2D CNN for further processing. PointPillars is used as the 3D object detector in the earlier version of the Autoware Foundation's open-source software for self-driving vehicles \cite{autowareauto}. 
    
Additionally, for a comprehensive evaluation of introspection in recent detection mechanisms, we employ the CenterPoint model \cite{centerpoint}. This model is distinct in its approach to object detection, as it focuses on identifying the center of objects first, and then regresses to define the bounding box. This is in contrast with other detectors that directly regress the corners of the bounding box. The fundamental motivation behind CenterPoint is the property that the centers of objects remain invariant to rotation. That ensures reliable detection even when vehicles assume different orientations due to varying road conditions. In addition, CenterPoint is the model utilised in Autoware Foundation's never versions, ``Autoware.Universe"~\cite{autowareuniverse}. In terms of implementation and training of these models, we have employed OpenMMLab's OpenMMDet3D framework \cite{mmdet3d} and utilised the pre-trained models on Kitti~\cite{kitti} and NuScenes~\cite{nuscenes} datasets. In this framework, both models utilise a network called SECOND~\cite{SECOND}, which applies sparse convolution operations to provide faster operation with the sparsity of the LiDAR's point cloud data.

\subsection{Datasets}
To measure the performance of introspection mechanisms, we utilise two widely-used datasets based on their use in both introspection and ADS domains: Kitti~\cite{kitti} and NuScenes~\cite{nuscenes}. The Kitti dataset consists of over 14,000 annotated images captured by a camera and a Velodyne LiDAR mounted on a car driving through urban environments in Karlsruhe, Germany. The training set contains 7,481 annotated samples,while test set includes 7,518. The benchmarking is typically done using only three classes: car, pedestrian, and cyclist. Additionally, the data labelling only covers objects in front of the vehicle.

The NuScenes dataset contains 1,000 diverse driving scenes captured across various urban locations. It contains, featuring multiple camera feeds, RADAR, and a full 360-degree LiDAR. In total, the NuScenes dataset includes 1.4 million images, 390k LiDAR sweeps, and 1.4 million 3D bounding box annotations across 23 object classes. The dataset is split into a training set, a validation set, and a test set. Additionally, the dataset provides detailed annotations not just for objects in front of the vehicle but in its entire surroundings, offering a 360-degree perspective. 

\subsection{Introspection Mechanisms}

For comparison purposes, we adapt the operation of two SOTA introspection mechanisms for 2D object detection, which demonstrated strong performance in the field~\cite{rahmanper, hakaniccv}, to become capable of handling 3D point cloud data. Each method applies a distinct processing technique on the raw activation layers, as explained below.  

First, in~\cite{rahmanper}, the authors utilised the activation maps of the object detector's backbone CNN to extract features for error detection. In their implementation, the activations are extracted from the last layer of a pre-trained ResNet50 and provide a comprehensive view of the learned features that can assist in interpreting the neural network's response to the input image. The authors apply mean, max, and standard deviation operations on the activation maps, which are originally 3D, $\text{height}\times \text{width} \times \text{channels}$ (or $H\times W\times C$). To generate and convert the learning representation into a 1D vector, global pooling is applied across height and width. The resulting vectors are concatenated into a column vector, which is used to train a multi-layer perceptron for binary classification, i.e., `Error' or `No-Error'. 

Another study on the introspection of 2D object detection~\cite{hakaniccv} has recently shown promising results on Kitti~\cite{kitti} and Berkeley Deep Drive~\cite{bdd100k} datasets by adapting an out-of-distribution detection mechanism, which was originally proposed for image classification problems~\cite{ash}. Similar to~\cite{rahmanper}, the authors extracted the last layer activations from the backbone network and set the activations whose values are less than the p-th percentile of the activation map equal to zero. In their study, the best percentile value was empirically determined. However, it is likely that this mechanism won't alter any of the activations in LiDAR-based 3D object detection due to the sparsity of the point cloud data and its propagation on neural activation patterns. We have noticed that the non-zero activation values constitute much less than 1~\% of the activation map. Hence, this method is essentially no different than directly using last layer activations (LLA) for the introspection of LiDAR-based object detection.

The main goal of this study is to validate the efficacy of using activations from earlier layers for introspection. To this end, we extract the outputs from the point cloud processor module employed in both CenterPoint and PointPillars models. 
This module processes the point cloud data, transforming it into pseudo-images, which we hereafter call processed point cloud (PPC), see at the top of Figure~\ref{fig:proposedframework}. Besides that, we also focus on the middle layer from the backbone activation for introspection analysis. Due to implementation variations in OpenMMDet3D, the architecture of the SECOND network in PointPillars and CenterPoint differs significantly. In PointPillars, the SECOND network consists of three blocks, each comprising consecutive convolution, batch normalisation, and RELU layers. In contrast, CenterPoint omits the first block present in PointPillars. Consequently, for mid-layer activation extraction, we use the outputs from the second block in PointPillars and the first block in CenterPoint. Notably, both blocks yield 128 distinct activation maps. This approach allows us to compare the models on a similar basis, despite their structural differences and provides insights into the impact of mid-layer activations (MLA) on the introspection performance. Finally, we will also evaluate the performance of an introspection mechanism that leverages the PPC, MLA and LLA after spatial adjustments and concatenation, see Figure~\ref{fig:proposedframework}.

\subsection{Introspection Training and Implementation}
To generate the error datasets, we perform object detection using Kitti and NuScenes datasets with their corresponding 3D object detection model. In this process, we extract the activation maps and generate the error labels. To label a sample as an error, we go through each ground truth object and check if there is no predicted bounding box that has an intersection over a union greater than 0.7 with the ground truth object bounding box.

To train the introspection network, we have utilised stochastic gradient descent (SGD) optimiser with focal loss function \cite{focal}. Due to the imbalance in the error datasets, we have calculated class weights~\cite{classweight} using training data and fed to the loss calculation along with a gamma ($\gamma$) value of five to mitigate the issue. We also implemented an early stop mechanism with a patience setting of 15 epochs, coupled with a learning rate scheduler that scales down the learning rate by a factor of 0.7 after a patience period of 10 epochs. All networks were trained for a total of 200 epochs using this approach. We have experimented with learning rates of 0.01, 0.001, and 0.0005, among which 0.01 yielded the best performance. The batch size for this training was set at 64. Additionally, for neural network development and training, PyTorch and Torchvision were utilised. Detection evaluation and metric calculation is done with Torchmetrics~\cite{torchmetrics}. Complexity calculation and inference time calculation are done with thop and time library in Python. Finally, all experiments are done on a machine equipped with an Intel(R) Core(TM) i9-10980XE CPU and NVIDIA RTX 3090 GPU.

\subsection{Performance Metrics}
Since the design of the introspection method is based on a binary classification output for all models considered in this paper, the following metrics are selected to evaluate the performance. 
\begin{itemize}
\item \textbf{Area Under Receiver Operating Characteristic Curve (AUROC):} It provides an indicator of how well a classifier distinguishes between the positive (`error') and negative (`no-error') classes. It measures the model's ability to avoid false classifications, with a higher AUROC indicating better performance.

\item \textbf{Recall (Positive and Negative):} It measures the classifier's ability to correctly identify true positives and true negatives. Positive Recall (also known as \textit{Sensitivity} or True Positive Rate) quantifies the proportion of actual positives correctly identified by the model. Negative Recall (also known as \textit{Specificity} or True Negative Rate) quantifies the proportion of actual negatives that are correctly identified. High recall values for both positive and negative classes indicate a model's effectiveness in correctly classifying both error and no-error instances.
\end{itemize}

\subsection{Performance Comparison}\label{sec:perfcomp}
\begin{table}[t]
    \centering
    \begin{tabular}{ccccc}
    \toprule
         \makecell{\textbf{Dataset} /\\\textbf{Model}}&\textbf{ Input}&\textbf{Rec.$_{(-)}$}&\textbf{Rec.$_{(+)}$}&\textbf{AUROC} \\\midrule
         \multirow{5}{*}{\makecell{Kitti / \\ PointPillars} }&\makecell{SF}&0.1479&0.9408&0.6000\\
         &\makecell{PPC}&\textbf{0.7764}&\textbf{0.7524}&\textbf{0.8420} \\
         &\makecell{MLA}&\underline{0.7500}&\underline{0.7460}&\underline{0.8368} \\
          &\makecell{LLA}&0.6268&0.8105&0.8036 \\
         & Proposed&0.7077&0.7858&0.8309\\ \midrule
         \multirow{5}{*}{\makecell{NuScenes / \\ CenterPoint}} &\makecell{SF}&0.2607&0.9217&0.7322\\ 
         &\makecell{PPC}&0.7945&0.8995&0.9198\\
          & \makecell{MLA}&\textbf{0.7945}&\textbf{0.9060}&\textbf{0.9330}\\
          & \makecell{LLA}&0.7123&0.8581&0.8919\\
          & Proposed&\underline{0.8650}&\underline{0.8630}&\underline{0.9288}\\ \bottomrule
    \end{tabular}
    \caption{Error detection performance of introspection models including processed point cloud (PPC), middle-layer activations (MLA), last-layer activations (LLA), proposed concatenation, and statistical features (SF), on Kitti (using PointPillars detector) and NuScenes (using CenterPoint detector). Metrics include Recall for negative (Rec.${(-)}$) and positive classes (Rec.${(+)}$), and AUROC for overall classification capability. The best-performing model based on AUROC is highlighted in bold, and the second-best is underlined.}
    \label{tab:comparison}
\end{table}
In this section, we present a thorough evaluation of error detection mechanisms for 3D object detection in ADS encompassing (i) the proposed introspection method jointly leveraging the processed point cloud (PPC), mid-layer activation patterns (MLA) and last layer activation patterns (LLA), (ii) an introspection model using either the PPC, or activation patterns from the mid-layer or the last layer, and (iii) the method based on statistical features (SF). 
We explore how activation patterns in different layers influence the model's confidence and the error detection efficiency providing useful insights on the quality of different learning representations for introspection in 3D object detection. Activation maps are also presented for qualitative analysis. Finally, we assess the practicality of these mechanisms in real-world ADS through a computational complexity analysis confirming their feasibility for real-time applications.

\subsubsection{Detection Performance}
As presented on \Cref{tab:comparison}, the proposed model provides a competitive result in all metrics compared to the ones using earlier layer activations in Kitti dataset. Unlike other models, where more balanced performance is presented, our proposed model has a tendency around positive class reducing false negative rate. On the other hand, in the NuScenes dataset, the proposed model provides a more balanced result while maintaining overall performance, indicating its adaptability and effectiveness in a more complex driving dataset.
%

Alternatively, the earlier layers, PPC and MLA, show promising and competitive results. In the Kitti dataset, PPC shows a preference for negative class detection, while MLA demonstrates a slightly more balanced approach with second-best AUROC. We also see our model provides competitive performance with a tendency to positive class detection. 
In the NuScenes dataset, both PPC and MLA perform well with MLA having highest AUROC, but still doesn't match the balanced efficiency of the proposed model. Also, it is evident that both PPC and MLA outperform LLA which is in line with our hypothesis on this study.

Lastly, the LLA and SF models exhibit disparities in their performance. LLA shows moderate effectiveness, but its lower recall for the negative class in both datasets indicates a potential tendency in detecting `error' cases. This trait is accentuated in the SF model whose very low recall for the negative class and lowest AUROC, especially in the Kitti dataset, suggests a model that is highly skewed towards positive class detection, potentially at the expense of overall predictive accuracy.

\subsubsection{Model Confidence}

\begin{figure}[t]
    \centering
    \begin{subfigure}[b]{\linewidth}
        \includegraphics[width=\linewidth]{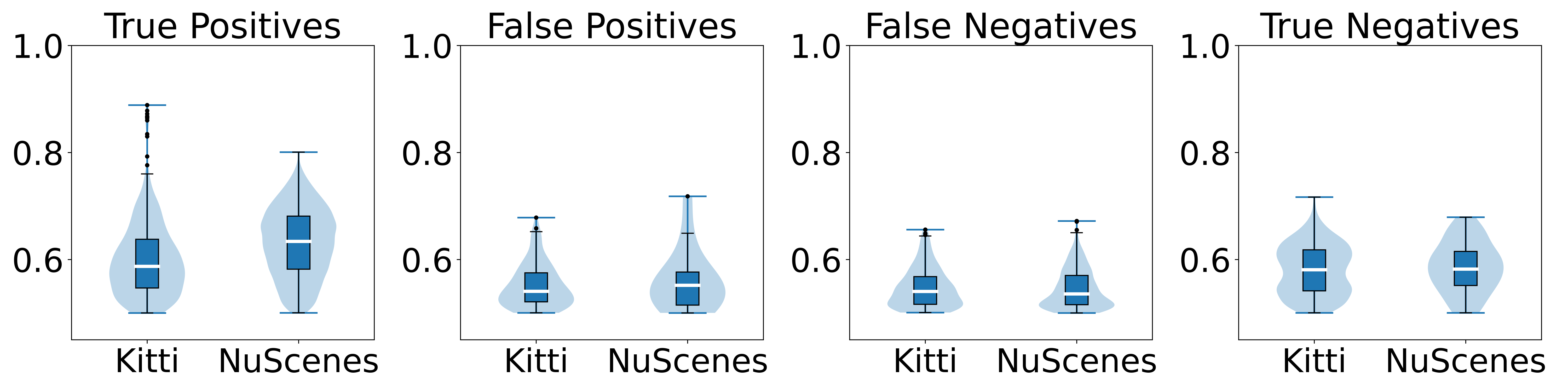}
        \caption{Processed Point Cloud (PPC)}
    \end{subfigure}
    \newline
    \begin{subfigure}[b]{\linewidth}
        \includegraphics[width=\linewidth]{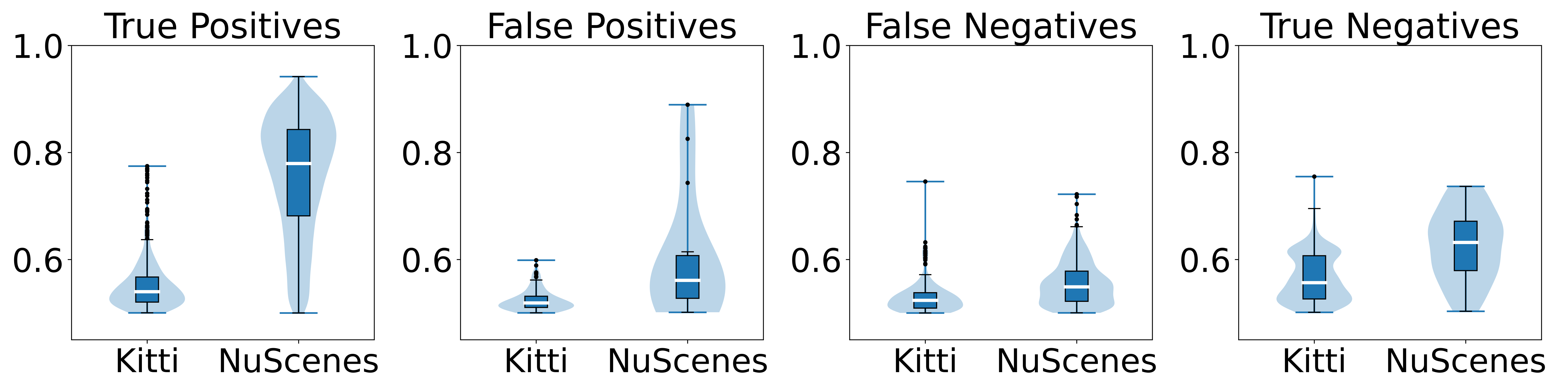}
        \caption{Middle Layer Activations (MLA)}
    \end{subfigure}
    \newline
    \begin{subfigure}[b]{\linewidth}
        \includegraphics[width=\linewidth]{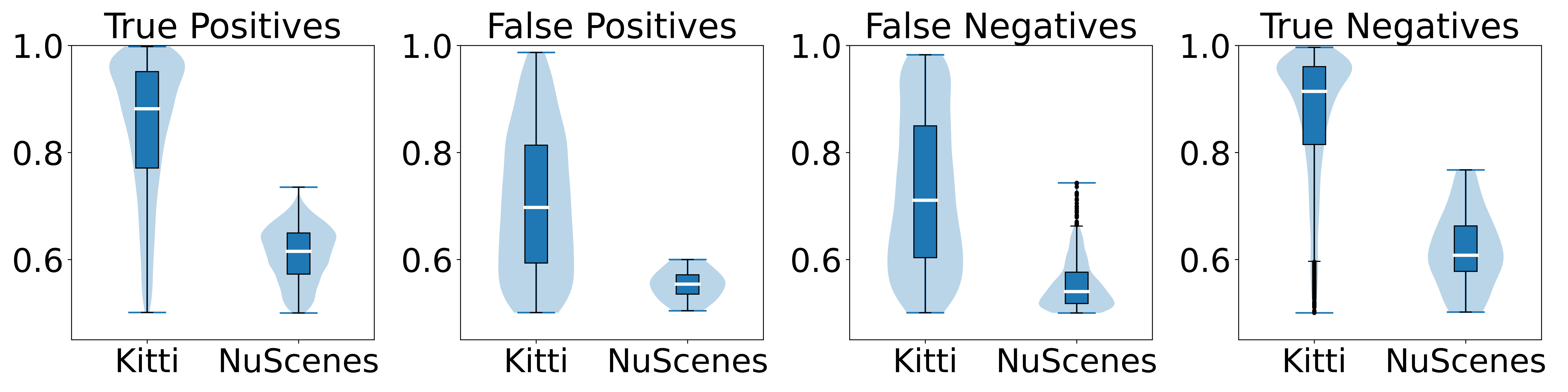}
        \caption{Last Layer Activations (LLA)}
    \end{subfigure}
    \newline
    \begin{subfigure}[b]{\linewidth}
        \includegraphics[width=\linewidth]{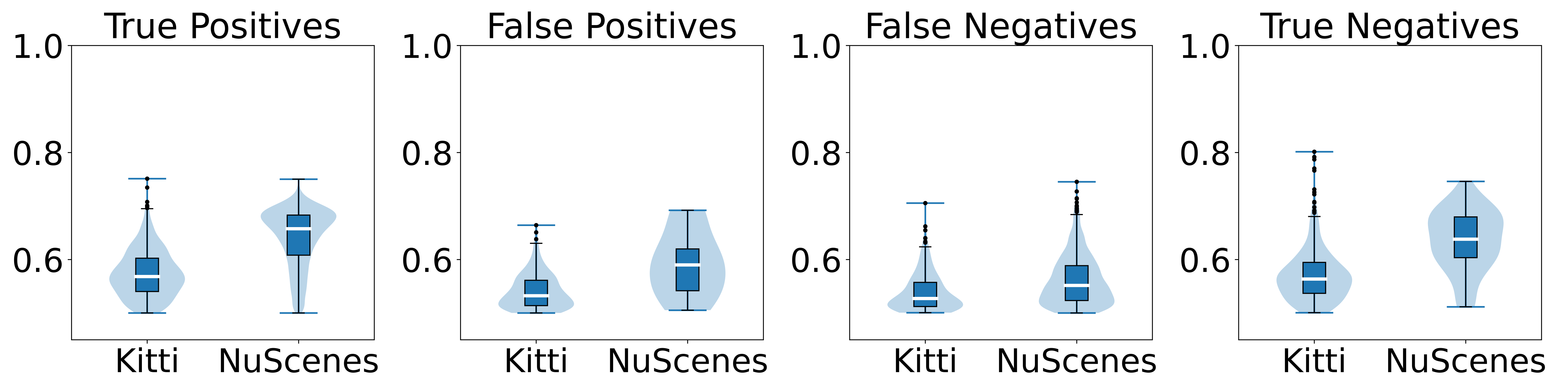}
        \caption{Proposed}
    \end{subfigure}
    
    \caption{Comparative analysis of confidence distributions across different inputs for introspection. The violin plots merged with a boxplot depict the confidence distributions for a trained neural network when tested on two datasets: Kitti and NuScenes. Each row represents an input modality:   (a) PPC, (b) MLA, (c) LLA, and (d) the proposed method. Within each modality, distributions are provided for true positives, false positives, false negatives, and true negatives. The width of each plot indicates the probability density of the data at different confidence levels, with mean and interquartile ranges also shown.}
    \label{fig:confidences}
\end{figure}

\Cref{fig:confidences} provides a comparative analysis of the confidence distributions for four distinct cases evaluated against the two datasets, Kitti and NuScenes. Specifically, the data is organised into four columns/categories — True Positives (TP), False Positives (FP), False Negatives (FN), and True Negatives(TN) — with confidence values (softmax outputs) that range between 0.5 and 1.
One may see that the distribution of confidence scores varies between the datasets, indicating that each dataset has unique learned characteristics that influence the introspector's confidence.

The PPC mechanism reveals  higher median confidence for TN and TP than FP and FN cases in both datasets, while the distinction is more apparent on Kitti dataset. This suggests that, in Kitti, this mechanism may perform better compared to other mechanisms, which is in accordance with the results depicted in \Cref{tab:comparison}. Additionally, we see the introspection is more confident for TP compared to TN in NuScenes dataset which is also reflected in \Cref{tab:comparison} indicating that the model is correct with TP and TN predictions.

The MLA mechanism demonstrates consistent high confidence in both TPs and TNs in NuScenes, where it performs the best according to \Cref{tab:comparison}. For FP and FN, although the ranges are wide, the majority of the predictions attain low confidence suggesting a dependable performance in correctly identifying positive and negative outcomes. In the Kitti dataset, while showing a similar pattern, the confidence intervals are narrower and predominantly skewed towards lower confidence values.

The LLA mechanism stands out with its very high confidence scores for almost all categories in Kitti. Yet, we see skewness towards higher values in TN and TP, which may indicate that the LLA mechanism has a strong ability to correctly identify both positive and negative outcomes with high confidence. However, considering the high confidence ranges in FP and FN, LLA could be prone to uncertainty. It is also evident that in more diverse and complex driving scenarios (NuScenes dataset), LLA mechanism is less confident in all categories.

The proposed mechanism shows confidence values up to 0.8 for TPs and TNs in the NuScenes dataset, implying that the network is generally reliable in its correct predictions but not overly confident. A similar pattern is also shown for Kitti, yet the model is not very confident for the majority of its decisions regardless of their correctness, yet it has higher upper confidence bounds for the correctly classified samples. Although the confidence distributions provide further insight about introspection, it is essential to also evaluate the model confidence with adversarial attacks and out-of-distribution samples in the future.
\subsubsection{Computational Complexity}

In this section, the computational requirements of each mechanism, focusing on the inference time and the number of floating point operations (FLOPs) metrics are considered. For this purpose, inference times of each model both on CPU and GPU, along with the FLOPs values, are provided for computational complexity comparison in \Cref{tab:timecomp}.
The `Proposed' method exhibits the lowest inference time on the CPU and a close second on GPU, significantly surpassing the performance requirements with GPU times under 2ms. The `PPC' method, while still under the threshold of 100ms, is considerably slower on the CPU, which may be critical in CPU-dependent scenarios. The reason for the slower inference is the higher resolution of earlier layers. On the other hand, the `MLA' and `LLA' methods offer a balanced trade-off between inference speed and FLOPs. In terms of FLOPs values, we see a similar trend with PPC being the highest, and again due to the resolution of the activation maps we have the lowest with LLA. As presented in the \Cref{tab:timecomp}, the proposed mechanism offers a significant reduction compared to the `PPC' mechanism and a slight improvement compared to `MLA' while preserving the error detection performance characteristics presented in \Cref{tab:comparison}. 
\begin{table}[t]
    \centering
    \begin{tabular}{cccc}
    \toprule
         \textbf{Method}&\makecell{\textbf{CPU}\\\textbf{Time (ms)}}&\makecell{\textbf{GPU}\\\textbf{Time (ms)}}
         & \makecell{\textbf{FLOPs (G)}} \\ \midrule  
         PPC &54.32 (9.54) &11.47 (1.21)&36.32\\
         MLA &9.43 (3.26) & 2.01 (0.10) & 3.68\\
         LLA &5.01 (0.47) &1.80 (0.06) &1.60 \\ 
         Proposed&4.94 (0.32) &1.95 (0.07)&2.60  \\\bottomrule
    \end{tabular}
    \caption{Average inference time with associated standard deviation (in parentheses) and floating point operations (FLOPs) for each introspection method using the Kitti dataset. The statistics are calculated based on 1000 iterations excluding initial warm-up (700-800 ms), on an Intel(R) Core(TM) i9-10980XE CPU and NVIDIA RTX 3090 GPU setup. The time-lapse is measured from the point where the backbone network outputs all activation patterns till the point where the introspection model provides its output.}
    \label{tab:timecomp}
\end{table}

\subsubsection{Qualitative Comparison}

\begin{figure}[t]
    \centering
    \includegraphics[width=\linewidth]{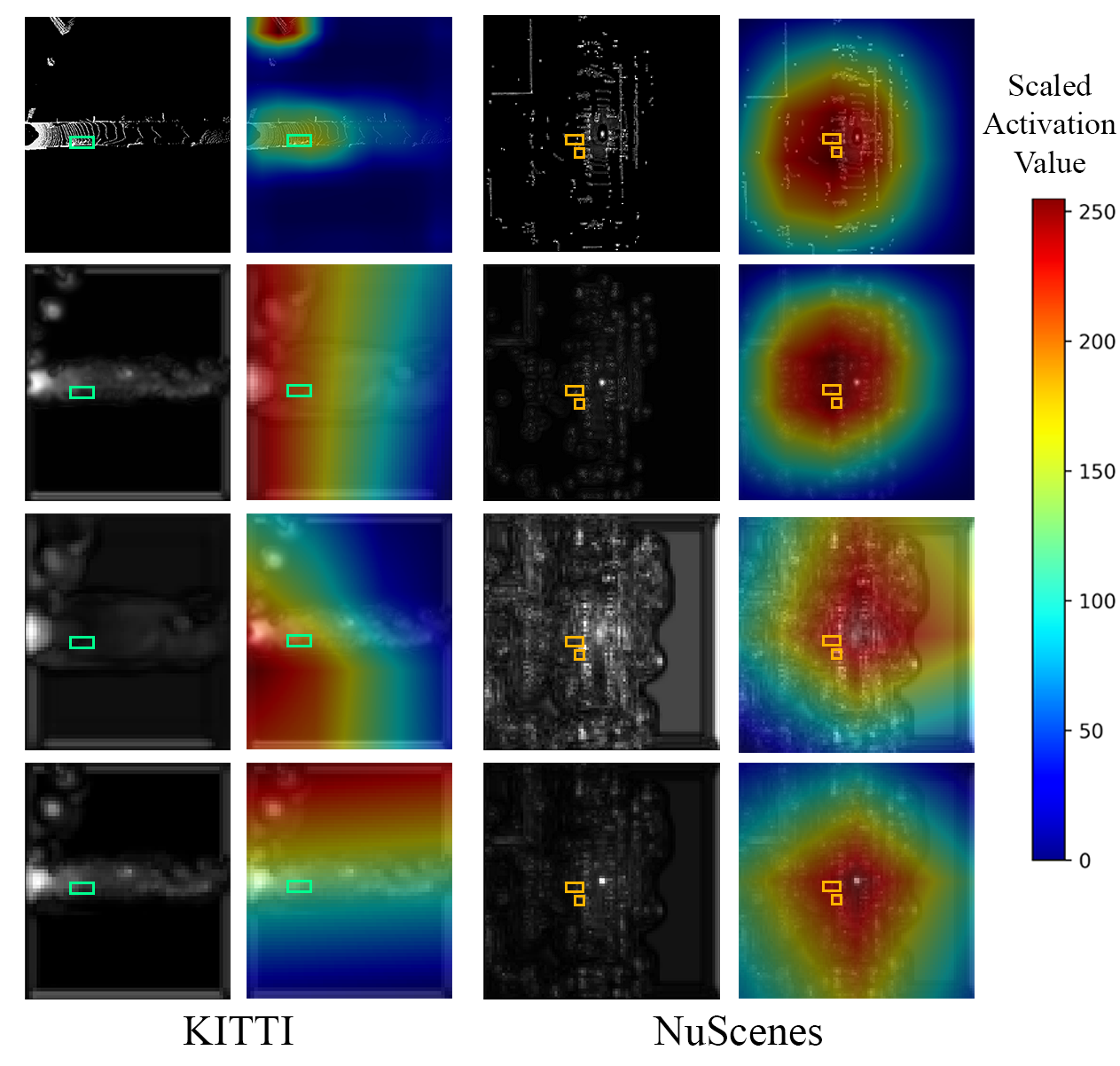}
    \caption{Max activation maps and Eigen-CAM visualisations for example frames on Kitti and NuScenes datasets. Every row represents a different activation map modality: PPC, MLA, LLA, and proposed. The first and third columns display the channel-wise max activations for the Kitti and NuScenes datasets, while the second and fourth columns exhibit the respective Eigen-CAM heatmaps that highlight areas critical to the classification. For clarity, objects correctly detected are marked with green boxes, while missed ones are highlighted with orange boxes.}
    \label{fig:activis}
\end{figure}

This section aims to provide an intuitive understanding of the effects that the activation patterns have on the decision-making of the introspection models considered in this paper. 
For this purpose, we have extracted the last layer activation maps of the introspector's CNN, i.e. ResNet 18, and applied well-known class-activation map generation method, Eigen-CAM \cite{eigencam}, which simply calculates the principal components of the activation maps and generates visualisations highlighting the activated regions. The resulting visualisations are presented for each mechanism and for both datasets in \Cref{fig:activis}. The driving direction is from left to right in Kitti and from bottom to top in NuScenes. It is also important to recall the distinction between the two  datasets: While Kitti focuses on objects in front of the vehicle, NuScenes provides a comprehensive 360-degree view. 

In the Kitti dataset (first and second columns), a no-error instance is showcased, exemplifying the precise detection of a vehicle situated ahead of the ego vehicle. For the introspection model using PPC input, the corresponding activation maps predominantly emphasise the road and the vicinity proximal to the ego vehicle, suggesting optimal initial layer performance in accordance with the results presented in \Cref{tab:comparison}. However, the maximum activation occurs at the top-left of the PPC input, where no relevant object exists, potentially indicating a case with reduced model confidence (\Cref{fig:confidences}). Moreover, the introspection models using activations from the middle and final layers, as well as the proposed introspection model, show a progressive, unilateral shift, with the MLA and LLA mechanisms concentrating on the ego vehicle's front and the proposed model concentrating on the left, yet maintaining road focus.

In the NuScenes dataset (third and fourth columns), an error case is given with missed detections of a pedestrian and a vehicle behind the ego vehicle which all introspection models correctly identify. The activation maps show concentrated activations around the ego vehicle, with the highest activation at the center. 
The left side of the ego vehicle receives more attention for PPC and MLA inputs, while a broader central area is activated for the LLA and the proposed methods. Similar to Kitti, the highest activation areas in PPC, MLA, and the proposed methods correlate with the location of objects. However, performance wise, the best performing models are the MLA and the proposed according to  \Cref{tab:comparison}.


\section{Summary \& Conclusions}\label{sec:conc}
In this research, we investigated the impact of earlier and concatenated layers of neural activation patterns on the error detection performance of 3D object detection in automated driving systems (ADS). 
We hypothesised that in the context of point-cloud data, characteristic of 3D environments, early layers can enhance the error detection capabilities. To test this, we employed PointPillars and CenterPoint to extract activations from various network stages and create an error dataset, focusing on the identification of false negatives to enhance safety and trust in ADS. We then trained a separate neural network on the error dataset using either the early layer activations or a combination of activations.

Our findings reveal that using early layer neural activation patterns enhances the error detection capability in 3D object detection, as compared to using only the last layer activations, at the cost of processing time and computational resources owing to the higher resolution. 
Combining activations from multiple layers into the introspection framework offers a more balanced approach in terms of performance and complexity. In addition, it empowers the introspection model with the capability to successfully identify object detection errors without raising unnecessary alerts, which is paramount for ADS. Given that introspection in 3D object detection, particularly in ADS, is a relatively unexplored subject, further research is imperative. Future studies should focus on developing metrics for constructing error datasets, evaluating introspection performance in various 3D object detection applications, and assessing the domain-shift capabilities of introspection mechanisms. Moreover, more sophisticated methods for utilising activation patterns from multiple neural network layers should be explored. 

\section*{Acknowledgement}
This research has been conducted as part of the EVENTS project, which is funded by the European Union, under grant agreement No 101069614. Views and opinions expressed are however those of the author(s) only and do not necessarily reflect those of the European Union or European Commission. Neither the European Union nor the granting authority can be held responsible for them. This research has been also supported by the Centre for Doctoral Training (CDT) to Advance the Deployment of Future Mobility Technologies at the University of Warwick.
{
    \small
    \bibliographystyle{ieeenat_fullname}
    \bibliography{main}
}


\end{document}